\documentclass[sigconf]{acmart}

\usepackage{booktabs} 

\copyrightyear{2020}
\acmYear{2020}
\setcopyright{acmlicensed}\acmConference[GECCO '20 Companion]{Genetic and Evolutionary Computation Conference Companion}{July 8--12, 2020}{Cancún, Mexico}
\acmBooktitle{Genetic and Evolutionary Computation Conference Companion (GECCO '20 Companion), July 8--12, 2020, Cancún, Mexico}
\acmPrice{15.00}
\acmDOI{10.1145/3377929.3398095}
\acmISBN{978-1-4503-7127-8/20/07}

\begin{document}
\title{Real World Morphological Evolution is Feasible}

\author{T{\o}nnes F. Nygaard}
\affiliation{%
  \institution{University of Oslo, Norway}
}
\email{tonnesfn@ifi.uio.no}

\author{David Howard}
\affiliation{%
  \institution{CSIRO, Australia}
}
\email{david.howard@data61.csiro.au}

\author{Kyrre Glette}
\affiliation{%
  \institution{RITMO, University of Oslo, Norway}
}
\email{kyrrehg@ifi.uio.no}

\renewcommand{\shortauthors}{T.F. Nygaard et al.}


\begin{abstract}

Evolutionary algorithms offer great promise for the automatic design of robot bodies, tailoring them to specific environments or tasks. 
Most research is done on simplified models or virtual robots in physics simulators, which do not capture the natural noise and richness of the real world.
Very few of these virtual robots are built as physical robots, and the few that are will rarely be further improved in the actual environment they operate in, limiting the effectiveness of the automatic design process.
We utilize our shape-shifting quadruped robot, which allows us to optimize the design in its real-world environment.
The robot is able to change the length of its legs during operation, and is robust enough for complex experiments and tasks.
We have co-evolved control and morphology in several different scenarios, and have seen that the algorithm is able to exploit the dynamic morphology solely through real-world experiments.

\end{abstract}

\keywords{Evolutionary Robotics, Real-world Evolution}

\ccsdesc[500]{Computing methodologies~Evolutionary robotics}
\ccsdesc[300]{Theory of computation~Evolutionary algorithms}

\maketitle

\section{Introduction}

In this paper, we present our shape-shifting robot, and the work we have done evolving its body shape in real-world environments.
We show that evolutionary optimization is an effective technique for automated design, even when relying purely on real-world experiments on a physical robot.

The field of Evolutionary Robotics (ER) uses evolutionary computation techniques to improve various aspects of robots.
Most common is to optimize the control of a robot, either through higher level control policies like movement trajectories for wheeled or flying robots, or more lower level concepts like the gait of legged robots.
Most work is not done on physical robots, but on simplified models in a physics simulator~\cite{mouret201720}.
These are never 100\% accurate, and the discrepancy between behavior in simulation and in the real world is referred to as the reality gap~\cite{jakobi1995noise}.
It makes evolutionary methods less effective as their results are not tuned to the specific environment and task they are meant for, but a simplified and inaccurate model of it.
This is considered one of the biggest challenges in the ER field~\cite{eiben2014grand}.

\begin{figure}[t!]
    \includegraphics[scale=0.21]{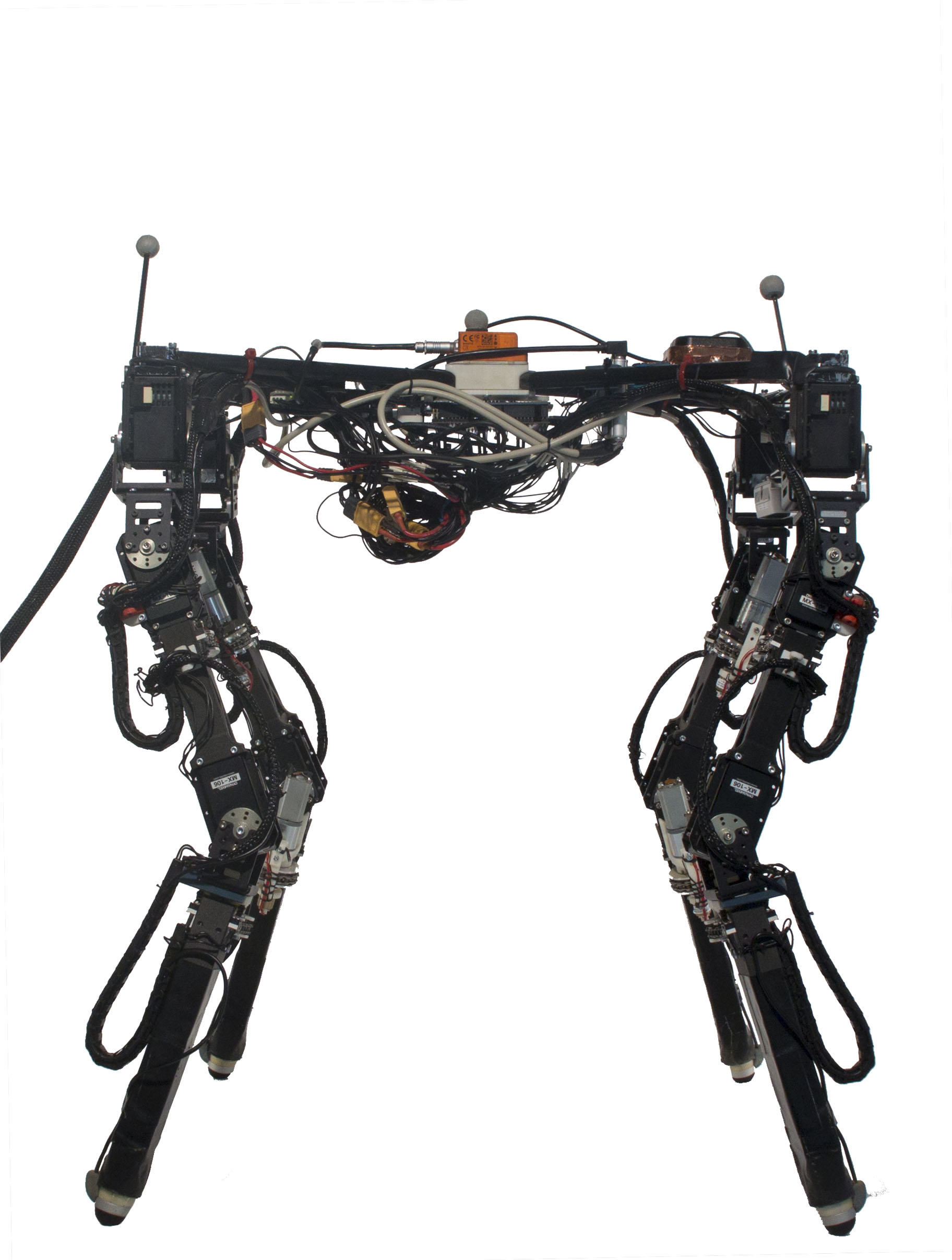}%
    \includegraphics[scale=0.21]{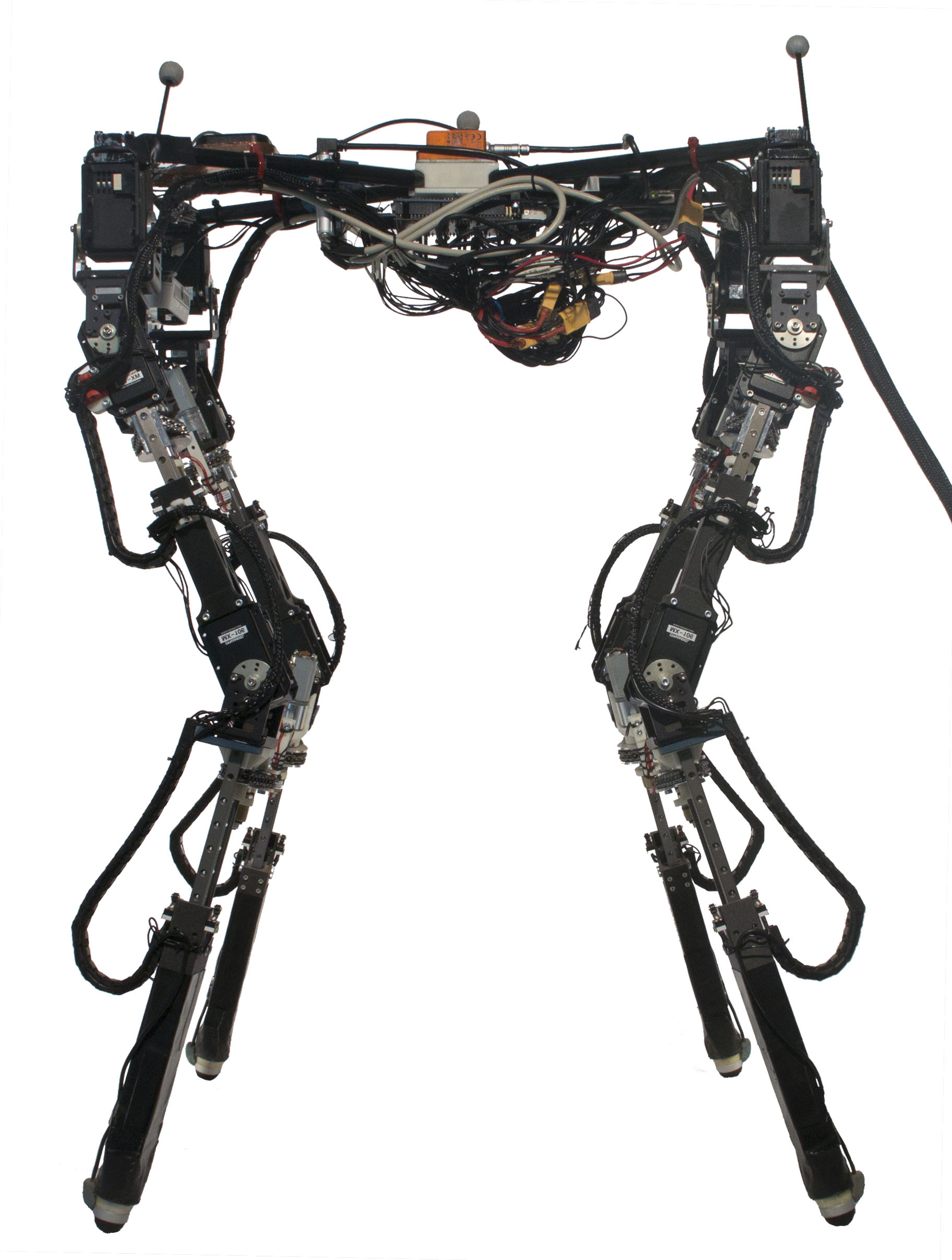}
    \caption{Our shape-shifting robot, with retracted legs to the left, and extended legs to the right.}
    \label{fig.lengthComparison}
    \vspace{-5mm}
\end{figure} 


There are many ways to reduce the impact of the reality gap when evolving control.
Firstly, the gap can be reduced significantly by using high-resolution simulators \cite{collins2018towards}.
Another common option is to do the start of the evolutionary run in simulation, before finishing the run in the real world~\cite{lipson2000automatic}.
This serves as a local search around the solutions found during simulation, which makes it less likely to find global optima only found in the real world.
Another approach is to try to improve the simulators using real-world data.
This can reduce the reality gap, but physics simulators or mathematical models will always be simplifications of the real world, and will thus never be perfect~\cite{mouret201720}.
With neural controllers, the use of plasticity for gap-crossing behavioural compensation has recently been shown possible~\cite{qiu2020crossing}, and using an embodied approach through sensor feedback has also shown some promise~\cite{nordmoen2019evolved}.
Avoiding parts of the search space where the reality gap is larger is also an option~\cite{koos2012transferability}.
The problem is that the best solutions are often the ones that are able to exploit the dynamics of the system and the peculiarities of the environment.
These are also the ones that are hardest to simulate and predict correctly, so this approach often removes most of the high performing individuals.

The only approach with the potential to completely bypass the reality-gap problem is doing evolution in the real world.
A challenge here, is that each evaluation can take a long time, and the physical robot(s) have to be designed, built, maintained and repaired, costing considerable amounts of effort~\cite{nygaard2019experiences}.
There are many examples where this has been done successfully, but these are often done in very simple environments in the lab~\cite{shen2012learning, hornby2000evolving, heijnen2017testbed}.
This does bypass the reality gap, but unless the robot is subjected to the real-world environment it will operate in, there will still be a discrepancy in behavior that will affect the performance.
Only a system that can continuously evolve during operation will not be subject to the challenges traditionally associated with the reality gap.

When evolving only the control of the robot, combining simulation and real-world experiments can be fairly straightforward.
The robot most often has the same freedom of control in the real world as in simulation. 
When it comes to the design, however, very few robots are able to change their body in a meaningful way to continue optimization in the real world after it has been built.
This means that a lot of work doing evolutionary design of robots only do the optimization in simulation, before optimizing control alone in the real world~\cite{lipson2000automatic, hornby2003generative, rosser2019sim2real}.
This is still an effective technique, but the reality gap lessens the potential efficiency and impact of the technique considerably~\cite{nygaard_WS_ICRA18}.
There has been a few examples of evolution of the body of the robot in the real world, but these require excessive time, human intervention, or the robots are very simple and limited. Broadly, the approaches seen to date have been (i) simulate some robots, then print them~\cite{collins2018towards}, or (ii) automatically combine predefined modules to create the phenotype~\cite{brodbeck2015morphological}.

\vspace{-2mm}

\section{Our approach}
Our main goal has been to do morphological optimization in the real world, and we wanted to do this on a functional robot design that could be used to solve real-world tasks.  The principal idea is that morphological adaptation is built into the robot, in this case enabled through extra motors built into the legs of the platform. 
Our choice fell on a four legged robot, which gives a good balance between power efficiency and control complexity requirements.
During the design and planning phase, we focused on developing a robust platform applicable to a range of use cases in machine learning and general robotics.

There are many different ways to parameterize the morphology of a four legged mammal-inspired robot.
We decided to pursue variable length legs, as this seemed technologically feasible with a good effect on behavior and performance.
The femur and tibia of each leg can all change their lengths by 50mm and 100mm respectively, as seen in Figure~\ref{fig.lengthComparison}.
We believe that changing the length of the legs allows the robot to do a mechanical gearing of the servos, which can trade speed for force, to adapt to situations where the trade-off between these vary.
We did initial experimentation both inside the lab and in realistic outdoor environments to investigate this effect, and showed that shorter legs worked better in challenging environments, while simple environments favored longer legs~\cite{tonnesfn19icra}.

Doing evolution exclusively in hardware has a range of associated challenges.
One of the most important, is that the evaluation budget is severely limited when compared to physics simulation.
The longer the experiments lasts, the higher the chance of sudden damage to the robot, as well as gradual wear and tear affecting the results.
Each evaluation can also take several seconds to minutes, and often requires human intervention throughout the experiment.
Having a system that is able to be adjusted to fit the evaluation budget at hand (as well as the requirements from the task and environment) is very important.
We have implemented a control system with variable complexity that can be adjusted to the evaluation budget, as well as the requirements from the environment and task~\cite{nygaard_evostar19}.

In our first evolutionary experiment, we wanted to investigate to what degree evolution would be able to exploit dynamic morphology, given a robot with adaptable body and control.
Changing control parameters gives a much more direct and commanding effect on the behavior of the system, so one might assume that the algorithm would spend most effort on control, and not necessarily utilize morphology to a significant degree.
We varied the voltage to our servos, resulting in two very different conditions for the robot.
Each evolutionary run took about an hour, and we did 6 runs in total.
Our experiments showed that evolution was able to adapt to the reduction in servo torque of about 20\%, and utilized both control and morphology to achieve this~\cite{tonnesfn_gecco18}.

We also wanted to investigate whether the robot would be able to exploit the dynamic morphology while walking on different surfaces.
We ran a longer evolutionary experiment on two different types of carpets, and saw that the evolutionary search yielded significantly different bodies for the two surfaces~\cite{nygaard2020environmental}.
Each full evolutionary run took just under two hours, and we did 10 runs in total.
We also tested the evolved individuals on previously unseen surfaces, where the robot performed better on surfaces qualitatively similar to the one they were evolved on, implying our method could be generalized to unknown terrains as well.

\begin{figure}
    \includegraphics[scale=0.052]{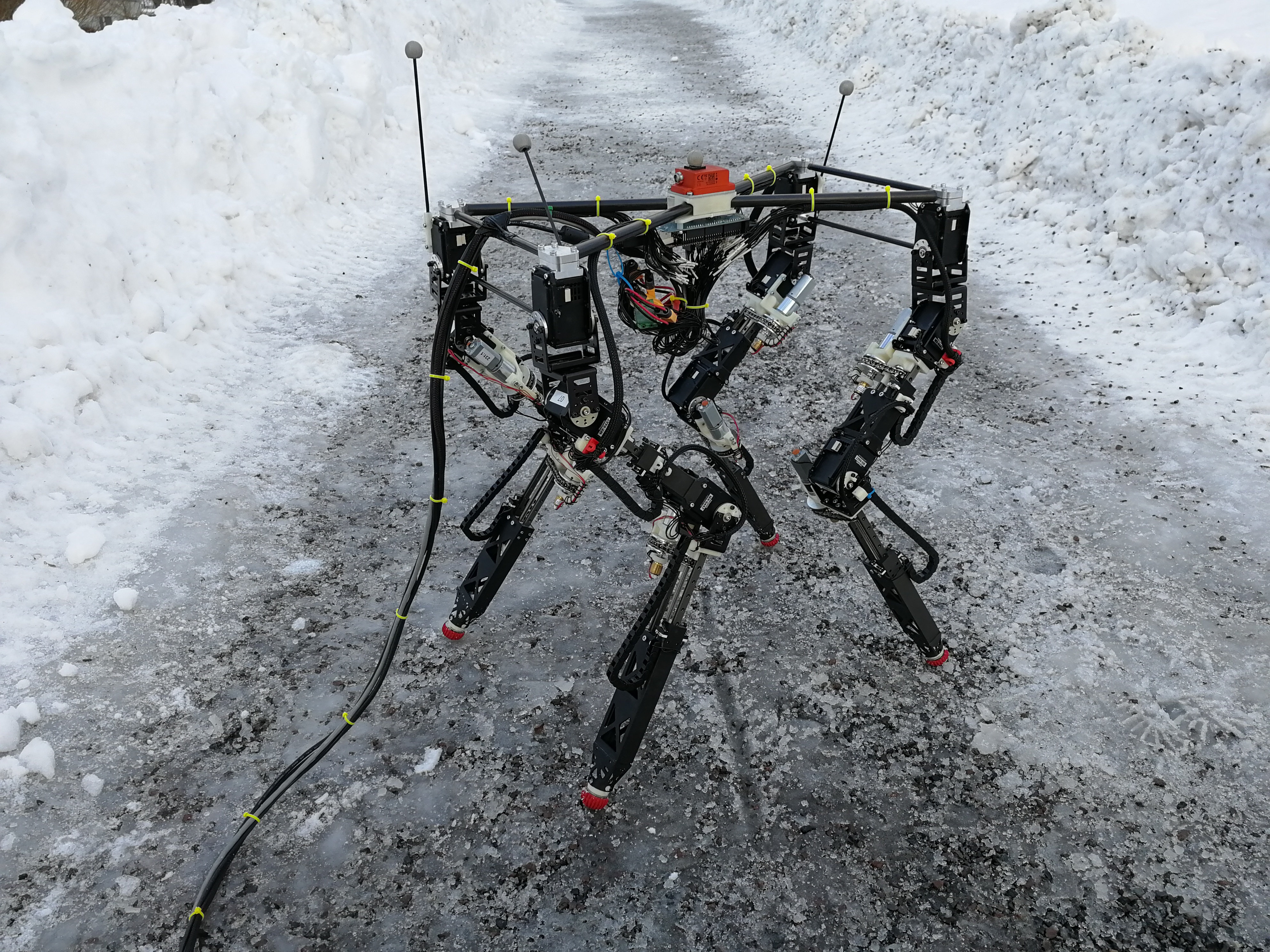}
    \caption{Our robot during testing in one of the outdoor environments, with ice and snow during Norwegian winter.}
    \label{fig.outside_path}
    \vspace{-5mm}
\end{figure} 

\vspace{-1mm}

\section{Conclusion}
In this paper, we have demonstrated the usefulness of real-world evolution of robot morphology and control -- a concept that has previously been considered unfeasible due to the obvious challenge of efficiently changing a robot's physical body.
By evolving in the real world, we show that evolution adapts both morphology and control to different real-world scenarios that would have been very hard --if not impossible-- to simulate.
We discuss the design considerations for our robotic platform making these kinds of experiments possible, and the level of abstraction necessary for the evolutionary optimization.
We hope to encourage further exploration of morphology and control evolution on real-world robots for adapting to complex environments.


\bibliographystyle{ACM-Reference-Format}
\bibliography{bibliography} 

\end{document}